\documentclass[10pt,twocolumn]{IEEEtran}
\usepackage[utf8]{inputenc}
\usepackage{geometry}
\usepackage{cite}
\usepackage{amsfonts,amsmath,bm}
\usepackage[ruled]{algorithm2e}
\usepackage{amsmath}
\usepackage{subcaption}
\usepackage{graphicx}
\graphicspath{{images/}}
\usepackage{pgfplots}
\usepackage{hyperref}
\usepackage{amsmath}
\usepackage{amssymb}

\pgfplotsset{width=0.45*\textwidth,compat=1.9}

\title{Deep Multi-Agent Reinforcement Learning for Decentralized Active Hypothesis Testing}
\author{{Hadar Szostak and Kobi Cohen}
	\thanks{
		H. Szostak and K. Cohen are with the School of Electrical and Computer Engineering, Ben-Gurion University of the Negev, Beer-Sheva, Israel (e-mail:hadarsz@post.bgu.ac.il; yakovsec@bgu.ac.il).} 
		\thanks{A short version of this paper was presented at the annual Allerton Conference on Communication, Control, and Computing (Allerton) 2022 \cite{szostak2022decentralized}.}
  \thanks{This work has been submitted to the IEEE for possible publication. Copyright may be transferred without notice, after which this version may no longer be accessible.}
	\thanks{This research was supported by the ISRAEL SCIENCE FOUNDATION
(grant No. 2640/20)}
	\vspace{-0.75cm}
}
\begin{document}

\maketitle
\begin{abstract}
\label{sec:abstract}
We consider a decentralized formulation of the active hypothesis testing (AHT) problem, where multiple agents gather noisy observations from the environment with the purpose of identifying the correct hypothesis. At each time step, agents have the option to select a sampling action. These different actions result in observations drawn from various distributions, each associated with a specific hypothesis. The agents collaborate to accomplish the task, where message exchanges between agents are allowed over a rate-limited communications channel. The objective is to devise a multi-agent policy that minimizes the Bayes risk. This risk comprises both the cost of sampling and the joint terminal cost incurred by the agents upon making a hypothesis declaration.

Deriving optimal structured policies for AHT problems is generally mathematically intractable, even in the context of a single agent. As a result, recent efforts have turned to deep learning methodologies to address these problems, which have exhibited significant success in single-agent learning scenarios. In this paper, we tackle the multi-agent AHT formulation by introducing a novel algorithm rooted in the framework of deep multi-agent reinforcement learning. This algorithm, named Multi-Agent Reinforcement Learning for AHT (MARLA), operates at each time step by having each agent map its state to an action (sampling rule or stopping rule) using a trained deep neural network with the goal of minimizing the Bayes risk. We present a comprehensive set of experimental results that effectively showcase the agents' ability to learn collaborative strategies and enhance performance using MARLA. Furthermore, we demonstrate the superiority of MARLA over single-agent learning approaches. Finally, we provide an open-source implementation of the MARLA framework, for the benefit of researchers and developers in related domains.
\vspace{2mm}

Index Terms - \textnormal{Active hypothesis testing (AHT), controlled sensing for multihypothesis testing, decentralized inference, deep reinforcement learning (DRL), multi-agent learning}
\end{abstract}

\section{Introduction}

We consider a decentralized sequential hypothesis testing problem, where multiple agents gather noisy observations from the environment in a sequential manner. Their goal is to determine whether the accumulated data provide evidence for a specific hypothesis. Notably, each agent's perspective might be limited, and they are capable of selecting a sampling action from their individual set of choices at each time step. These actions yield observations following distinct distributions based on the given hypothesis. Due to communication limitations, sharing the complete observation space among agents is not feasible. Nevertheless, message exchanges between agents are allowed over a rate-limited communication channel, enabling collaboration towards task completion. The primary objective is to devise a multi-agent policy that minimizes the Bayes risk. This risk encompasses both the expense of sampling and the collective final cost incurred by agents when they make their hypothesis declarations. This problem has implications in various multi-agent systems, including setups like multi-robot systems, multi-drone systems, and sensor networks. 

In the following subsections, we will explore how this problem aligns with prior research and formulations. We will also elaborate on how this research advances and extends classic theories in various aspects.

\subsection{Single-Agent Active Hypothesis Testing}

In contrast to fixed-sized hypothesis testing, which involves a predetermined sample size, sequential hypothesis testing represents a statistical approach pioneered by Wald \cite{wald1947sequential}. This methodology allows for a variable sample size that is not determined beforehand. In the context of sequential hypothesis testing, an agent accumulates observations in a sequential manner. The agent follows a predefined stopping rule, concluding the sampling and asserting a hypothesis once there is a dependable level of evidence to support a specific hypothesis. Wald introduced the sequential probability ratio test (SPRT), which stands as an optimal technique in terms of minimizing sample size while adhering to constraints related to type I and type II errors.

An extension of the sequential hypothesis testing discussed earlier is the sequential experiment design problem, initially explored by Chernoff in 1959 \cite{chernoff1959sequential}. In comparison to the classical Wald's sequential hypothesis testing, where the observation model for each hypothesis is predefined, sequential experiment design introduces an element of control. This empowers the agent to select the experiment to conduct at each time step. Different experiments yield observations following distinct distributions under each hypothesis. The concept is intuitive: As more observations are gathered, the agent's certainty regarding the true hypothesis increases, leading to more informed choices in experiments. The objective is to minimize the Bayes risk, which encompasses both the cost of sampling and the final cost incurred by a single agent when making a hypothesis declaration.

Chernoff focused on scenarios involving binary hypotheses and devised an algorithm known as the Chernoff test. This algorithm is asymptotically optimal in terms of minimizing the Bayes risk as the error probability approaches zero. Subsequent research explored variations and extensions of the problem \cite{bessler1960theory, nitinawarat2013controlled, nitinawarat2015controlled, naghshvar2013active, naghshvar2013sequentiality}. In particular, the problem was referred to as controlled sensing for hypothesis testing in \cite{nitinawarat2013controlled, nitinawarat2015controlled} and active hypothesis testing (AHT) in \cite{naghshvar2013active, naghshvar2013sequentiality}. Asymptotically optimal solutions to the problem have been developed in these studies. 

Given the general mathematical intractability of obtaining optimal solutions for AHT problems, recent suggestions have leaned towards employing methods based on deep reinforcement learning (DRL) to tackle single-agent AHT challenges. DRL algorithms have garnered significant attention due to their ability to offer effective approximations of objective values, particularly when confronted with extensive state and action spaces. A notable instance of DRL's prowess comes from DeepMind's seminal Nature paper \cite{mnih2015human}, which introduced a DRL algorithm capable of instructing computers to directly play Atari games from on-screen pixels, showcasing impressive performance across numerous games. Furthermore, in the work by \cite{foerster2016learning}, notable achievements were demonstrated in various players' performances across MNIST games and the switch riddle. A particularly intriguing approach, referred to as Double Q-learning, was presented in \cite{van2016deep}. This method employs two Deep Neural Networks (DNNs) to concurrently learn the policy and value evaluation. A good resource for delving into the realm of DRL research is provided by \cite{li2017deep}. DRL has been investigated in recent years to solve active statistical inference problems \cite{kartik2018policy, puzanov2018deep, zhong2019deep,  puzanov2020deep, joseph2020anomaly}. Furthermore, it allows efficient implementations in real-world tasks and cheap-hardware agent devices \cite{livne2020pops}. Specifically, DRL-based algorithms designed for solving single-agent AHT problems can be found in \cite{zhong2019deep, joseph2020anomaly, kartik2018policy}. These studies underscore the emerging role of DRL in addressing AHT challenges within a single-agent framework. In \cite{kartik2018policy}, a Q-learning method was used to design experiment selection strategy $g$ for the agent such that the confidence level on the true hypothesis increases as quickly as possible. In \cite{zhong2019deep,joseph2020anomaly}, Actor-Critic DRL algorithms were used.

Nevertheless, note that all of these aforementioned studies focused on the single-agent context of AHT. Consequently, they are unable to solve the multi-agent AHT setting, which is the focus of this paper. Within our study, we elevate the scope of the problem to encompass a scenario where multiple agents, each potentially equipped with only a partial perspective of the environment, gather noisy observations. The collective objective is to determine whether the amassed data provide substantial support for a specific hypothesis. In line with the principles of AHT, we grant the agents observation control, enabling them to opt for a sampling action at each time step. These distinct actions result in observations originating from varying distributions under each hypothesis. Central to this investigation is the collaborative aspect where agents must work together to achieve the objective. This collaboration is facilitated through message exchanges between agents, facilitated by a rate-limited communications channel. The fundamental challenge at hand revolves around the formulation of a multi-agent policy. This policy must be crafted in a way that facilitates effective collaboration among agents, ultimately leading to the minimization of the Bayes risk. This risk encompasses both the cost incurred by sampling and the joint terminal cost borne by agents upon hypothesis declaration.

\subsection{Multi-Agent Hypothesis Testing}

Another generalization of Wald's sequential hypothesis testing discussed in the previous subsection is the decentralized Wald problem pioneered by Teneketzis in 1987 \cite{teneketzis1987decentralized}. Teneketzis considered a similar setting of sequential sampling as in Wald's setting, where the observation model under each hypothesis is predetermined. The new aspect in Teneketzis' model is the decentralization of the problem, where two agents make decisions, and the Bayes risk consists of the sampling cost and the joint terminal cost among agents. Specifically, if both agents infer correctly about the true hypothesis, the terminal cost is zero. If only one agent is correct, the cost is $1$, and if both agents are wrong, then the cost is greater than $1$. It was shown that the optimal decision is made using coupled thresholds for the agents. The thresholds are calculated using a coupled Dynamic Programming (DP) equations, where Teneketzis offered an approximate solution to the DP equations. Subsequent studies considered variations and extensions of the problem (see \cite{nayyar2014signaling, cui2015computing} and subsequent studies). Obtaining structured optimal solutions remained open for the general problem models. 

In contrast to Teneketzis' decentralized Wald problem, which relies on predetermined observation models for each hypothesis, the formulation under consideration incorporates an element of control. This empowers each agent to make decisions about which experiments to conduct at each time, much like the classic (single-agent) AHT scenario. Consequently, we refer to this problem as the decentralized AHT problem.

The development of DRL-based algorithms for multi-agent learning in the context of AHT problems represents a relatively new research direction. This research direction originated from our short version of this paper, which was presented at Allerton conference, 2022 \cite{szostak2022decentralized}. In that condensed version, we introduced the problem, outlined the algorithm, and presented initial simulation results. In this comprehensive version, we present an intricate exposition and explanation of the algorithm, provide an open-source software implementation of the algorithm, present a more extensive set of simulation results, engage in a thorough discussion of these findings, and provide a comprehensive analysis and comparison with existing literature. An alternative multi-agent formulation was recently explored in \cite{stamatelis2023deep}, where the objective is to minimize the expected maximum detection time among all agents, along with terminal costs and cost constraint. However, this differs from our aim, which is to minimize the total cost (i.e., Bayes risk) \cite{teneketzis1987decentralized} comprises the cost of sampling and the joint terminal cost incurred by the agents upon making a hypothesis declaration. Furthermore, \cite{stamatelis2023deep} assessed performance when exact measurements were exchanged between agents, which can be communication-intensive in real-time applications. In contrast, our paper demonstrates effective collaborative learning among agents operating over rate-limited channels, where only concise messages specifying selected actions are shared, without sharing the observed data.

It should be noted that multi-agent learning has been studied in different domains using online learning and and deep learning, and can be broadly divided into two groups: Competitive and cooperative multi-agent learning (a good overview can be found in \cite{zhang2021multi}). In competitive multi-agent tasks, each agent (or group of agents) strives to maximize its individual reward, with little consideration for the performance of other agents. Conversely, cooperative multi-agent tasks involve agents working together to maximize a shared reward, which can also be coupled among the agents themselves. Furthermore, the reward structure for each agent might involve a trade-off between individual and social rewards, as observed in \cite{cohen2016distributed, paul2023multi}. The multi-agent setting can leverage the presence of multiple agents in the environment to improve the single agent performance. In our framework, we fall within the cooperative setting. Here, agents collaborate to achieve a collective goal, aligning with the objective of minimizing the Bayes risk and achieving superior performance through collaboration among multiple agents in the decentralized AHT problem.

\subsection{Main Results}

We will now summarize the main results of this paper:

\noindent
\textbf{1) Decentralized AHT formulation:} We investigate a novel model of the decentralized AHT problem. In this setup, multiple agents sequentially collect noisy observations from the environment and must determine whether the gathered data provide support for a particular hypothesis. Each agent's perspective might be limited, granting them the ability to choose a sampling action from their respective sets of actions at each time step. These distinct actions result in observations originating from different distributions under each hypothesis. Due to communication constraints, it is not feasible to share the complete observation space among the agents. However, they can engage in message exchanges over a rate-limited channel, allowing for collaboration in achieving the task. The primary objective is to obtain a multi-agent policy that minimizes the Bayes risk, encompassing both the cost incurred through sampling and the collective terminal cost when agents make their hypothesis declarations. Our investigation into the decentralized AHT problem represents a significant generalization and advancement of classical theories in various aspects. In contrast to the classic single-agent AHT problem, this paper addresses scenarios where multiple agents, each potentially having a limited view of the environment, gather noisy observations to infer the true hypothesis based on their observed data. As in the AHT context, agents have observation control, enabling them to select a sampling action at each time step, resulting in diverse observations under different hypotheses. Collaboration is a key element as agents must learn to work together effectively, utilizing message exchanges over a constrained communication channel, all while minimizing the Bayes risk, comprising both sampling and terminal costs. Distinguishing our formulation from Teneketzis' decentralized Wald problem, where observation models are predetermined under each hypothesis, our approach introduces a control aspect. This allows each agent to make real-time decisions about which experiments to conduct, akin to the classic single-agent AHT setting. Consequently, we term this problem the decentralized AHT problem.

\noindent
\textbf{2) Algorithm development:} We develop a novel algorithm based on deep multi-agent reinforcement learning framework to solve the decentralized AHT problem, dubbed Multi-Agent Reinforcement Learning for AHT (MARLA). Specifically, our approach involves agents learning a policy that maps a given state to an appropriate action. This policy is learned through online exploration of actions and observing the resulting rewards, all with the overarching objective of minimizing the Bayes risk. Our multi-agent framework includes a communication mechanism that permits limited message exchanges between agents. Due to communication constraints, the exchange of explicit observations between agents is typically avoided. In our experimental results, only concise messages specifying selected actions are shared. Consequently, agents must learn the most effective sampling strategy and decision rules regarding the true hypothesis based on their individual observations and the shared, rate-limited messages. The proposed MARLA algorithm stands as the first of its kind to solve this particular problem. MARLA leverages an Actor-Critic DRL method, utilizing Proximal Policy Optimization (PPO) to ensure stable updates. In this approach, each agent autonomously selects sampling actions and stopping actions in a distributed manner, informed by their own observation history and the information contained in the shared short-messages. 

\noindent 
\textbf{3) Open source software:} We developed an open source software implementation of the MARLA algorithm. MARLA was developed using Python and is available at GitHub (see link in \cite{szostak2023deep_code}). Providing an open-source deep learning frameworks for MARLA would allow implementation and adjustments of the framework for the benefit of other researchers and developers in related fields.

\noindent
\textbf{4) Experimental Study:} We conducted experiments using MARLA within a custom environment designed for active anomaly detection, a task that has received extensive attention in recent years across various settings and assumptions, as documented in the literature \cite{lai2011quickest, caromi2012fast, tajer2013quick, cohen2014optimal, cohen2015active, cohen2015asymptotically, geng2016quickest, heydari2016quickest, kaspi2017searching, huang2018active, heydari2018quickest,  tsopelakos2019sequential, gurevich2019sequential, rosenthal2020arba, hemo2020searching, wang2020information, gafni2021searching, lambez2021anomaly, tsopelakos2022sequential, chaudhuri2022joint, kartik2022fixed}. Specifically, the task is to locate an anomalous process among a set of processes. Normal processes, when sampled, yield noisy observations conforming to a typical distribution associated with normal behavior. On the other hand, an anomalous process, when sampled, produces noisy observations deviating from the typical distribution. Agents can influence the observation distribution by selecting which process to sample at any given moment. It is essential to emphasize that computing a structured optimal sampling strategy for this problem is mathematically intractable, even in the single-agent setting. As a result, only asymptotically optimal solutions have been developed in previous work \cite{chernoff1959sequential}. In this paper, we tackle an even more challenging multi-agent setting. We conducted comprehensive simulations to test MARLA within this environment. Our experiments showcase the agents' capability to learn how to collaborate effectively, leading to performance improvements. The results clearly demonstrate that MARLA is superior to single-agent learning.

\section{System Model and problem statement}

We start by delineating the system model in Subsection \ref{ssec:system}, followed by an elaboration of the problem statement in Subsection \ref{ssec:problem}.

\subsection{Description of System Model}
\label{ssec:system}

We consider a decentralized AHT problem implemented by $K$ agents. Let $\mathcal{H}\triangleq\{H_1,...,{H_M}\}$ denote a set of $M$ hypotheses, and let $H\in\mathcal{H}$ be the true hypothesis. The a priori probability that $H_m$ is true, i.e., $H=H_m$, is denoted by $p_m$, where $\sum_{m=1}^{M}p_m=1$. To avoid trivial solutions, it is assumed that $0<p_m<1$ for all $m$. Let $\mathcal{A}^k$ be the set of actions, including sampling actions that control the observation distribution of the sample, and a stopping action that any agent can take to finalize its test and declare the hypothesis (while other agents may continue the test). We define the stopping rule $\tau^k$ as the time when agent $k$ finalizes the test by declaring the hypothesis. Let $\delta^k\in\{1, 2, ..., M\}$ be a decision rule, where $\delta^k=m$ if the agent declares that $H_m$ is true. At each time $n$, each agent $k$ broadcasts a short-message $m^k(n)$ (which can be a null message as well) to all other agents. At each time $n\in\{1, 2, ...\}$, agent $k$ can take a sampling action $a^k(n)\in\mathcal{A}^k$ based on the history of the task it has (i.e., all its past sampled observations, received messages, and actions) and obtain (high-dimensional) observation $o^k(n)$. The collected observation $o^k(n)$ is an independent random variable drawn from probability density $p^{k, a^k(n)}_H$ under the true hypothesis $H$. The time series vector of sampling actions is denoted by $\textbf{a}^k\triangleq(a^k(n), n=1, 2, ...)$. A policy for agent $k$ is given by the tuple $\Gamma^k\triangleq(\tau^k, \delta^k, \textbf{a}^k)$. The multi-agent policy for all $K$ agents is thus given by $\Gamma\triangleq\left\{\Gamma^k\right\}_{k=1}^K$.  

\subsection{Problem Statement: The Decentralized Active Hypothesis Testing Problem} \label{ssec:problem}

Let $c$ be the sampling cost for each observation for each agent. Let $J(\delta^1, \delta^2, ..., \delta^K, H)$ be the terminal cost incurred by the final decisions $\delta^1, \delta^2, ..., \delta^K$ of the agents, where $H$ is the true hypothesis. The terminal cost is coupled in general, thus, $J(\delta^1, \delta^2, ..., \delta^K, H)\neq \sum_{i=1}^{K}J(\delta^i, H)$. Otherwise, the problem decomposes into $K$ independent classic AHT problems \cite{naghshvar2013sequentiality}. Furthermore, we assume that the terminal cost increases with the number of agents (say $i$) that declare $\delta^i\neq H$. In other words, an increased number of mistakes leads to higher terminal costs. The Bayes risk under multi-agent strategy $\Gamma$ when hypothesis $H$ is true is given by: 
\begin{equation}
\label{eq:risk_H1}
     R_H(\Gamma)\triangleq\textbf{E}_H\left\{\left(\sum_{i=1}^{k}{c\tau^i}\right)+J(\delta^1, \delta^2, ..., \delta^K, H)\right\},
 \end{equation}
where $\textbf{E}_H$ denotes the operator of expectation with
respect to the probability measure under hypothesis $H$. The average Bayes risk is given by: 
\begin{equation}
\label{eq:risk_H2}
     R(\Gamma)\triangleq
     \sum_{m=1}^{M}p_m R_{H_m}(\Gamma).
 \end{equation}
The objective of the decentralized AHT problem is to find a multi-agent policy $\Gamma$ that minimizes the average Bayes risk $R(\Gamma)$:
\begin{equation}
\label{eq:opt}
     \inf_{\Gamma} R(\Gamma).
 \end{equation}

\section{The Multi-Agent Reinforcement Learning for AHT (MARLA) Algorithm}
\label{sec:marla}

In this section, we present the proposed MARLA algorithm to solve (\ref{eq:opt}), based on DRL optimization. DRL has gained significant attention in recent years due to its ability to provide effective approximations of objective values by integrating DNNs with reinforcement learning. Within this framework, DNNs are employed to establish mappings from states to actions, especially in large-scale models, with the aim of maximizing the objective value. In the context of the MARLA framework we have developed, our goal is to train a DNN within a multi-agent setting. Each agent is trained to map its present observation and communication signals to sampling actions, stopping actions, and decision rules, all based on the trained DNN. This is done by maximizing the accumulated (discounted) reward (or the negative Bayes risk in (\ref{eq:opt})). To handle the complex multi-agent environment, MARLA uses an Actor-Critic DRL method to train the agents, with Proximal Policy Optimization (PPO) to ensure stable updates. In the subsequent subsections, we delve into the various components that constitute the MARLA algorithm.

\subsection{Proximal Policy Optimization (PPO) Updates for Decentralized AHT}
\label{ssec:PPO}

The MARLA algorithm leverages the Proximal Policy Optimization (PPO) method, as introduced in \cite{schulman2017proximal}. PPO is a policy gradient reinforcement learning method that refines a policy by alternating between two key steps: sampling data from the policy and performing multiple epochs of optimization on the collected data. What sets PPO apart from traditional policy gradient methods is its approach to policy updates. Instead of drastically altering the policy, PPO seeks to make incremental adjustments while keeping the new policy update in close proximity to the older one within the policy parameter space. This proximity is maintained by employing a variant of Trust Region Policy Optimization (TRPO), as detailed in \cite{schulman2015trust}. TRPO focuses on ensuring that the new policy remains near the older one in terms of action probability distributions, essentially preserving the output of the policy.

As PPO operates offline in batches through small buffer, we use a different time index to represent the time series updates, denoted by the subscript $t$. Additionally, since a shared network is trained for all agents, we remove the superscript $k$ from our notations. Specifically, let $\pi_\theta(a_t|s_t)$ represent the output probability distribution over actions given state $s_t$, and $\pi_{\theta_{old}}(a_t|s_t)$ be defined similarly in relation to the policy parameters vector before the update, denoted as $\theta_{old}$. In TRPO, the objective function aims to maximize subject to a constraint on the size of policy update, measured in terms of KL divergence:
\begin{equation}
\begin{array}{l}
\label{TRPO}
\max_\theta \hat{E}_t\left[\dfrac{\pi_\theta(\alpha_t|s_t)}{\pi_{\theta_{old}}(\alpha_t|s_t)}\hat{A}_t\right]\vspace{0.2cm}\\
\mbox{s.t.\;\;}
\hat{E}_t\left[ KL\left[\pi_{\theta_{old}}(\cdot|s_t),\pi_\theta(\cdot|s_t)\right]\right]\leq\eta\;,
\end{array}    
\end{equation}
where
\begin{equation}\begin{split} \label{advantage_gae}
\hat{A}_t=\eta_t&+(\gamma\lambda)\eta_{t+1}+...+(\gamma\lambda)^{T-t+1}\eta_{T-1},
\end{split}
\end{equation}
and 
\begin{equation}\begin{split} \label{td_error}
\eta_t=r_t+\gamma V(s_{t+1})-V(s_t). 
\end{split}
\end{equation}
Here, $t$ is the time index on $[0,T]$ within a given length-T samples, $\gamma$ is the discount factor, $\lambda$ is the generalized advantage estimation (GAE) factor, $r_t$ is the reward and $V$ is the Value network. The TRPO optimization can be written as penalty-based formulation:
\begin{equation} \small \label{PPO_function}
    \max_\theta \hat{E}_t\left[\dfrac{\pi_\theta(\alpha_t|s_t)}{\pi_{\theta_{old}}(\alpha_t|s_t)}\hat{A}_t-\beta KL[\pi_{\theta_{old}}(\cdot|s_t),\pi_\theta(\cdot|s_t)]\right].
\end{equation}

TRPO employs a hard constraint because selecting a single penalty coefficient that performs well across various problems or even within a single problem with evolving characteristics can be challenging. To use a penalty-based formulation and overcome the problem of penalty coefficient selection, PPO introduces two solutions: Clipped Surrogate Objective and Adaptive KL Penalty Coefficient. While TRPO maximizes a Conservative Policy Iteration (CPI) \cite{kakade2002approximately} objective, i.e. Surrogate objective, we have:
\begin{equation} \label{CPI}
    L^{CPI}(\theta)=\hat{E}_t\left[r_t(\theta)\hat{A}_t\right],
\end{equation}
where 
\begin{equation} \label{new_policy_old_policy}
    r_t(\theta)=\dfrac{\pi_\theta(\alpha_t|s_t)}{\pi_{\theta_{old}}(\alpha_t|s_t)}.
\end{equation}

In \emph{Clipped Surrogate Objective} (referred to as CLIP in the notations), we apply clipping to changes in the policy that cause the probability ratio between the new and old policy distributions to deviate from 1:
\begin{equation} \small \label{clipping}
\begin{split}
    L^{CLIP}&(\theta)=\\&\hat{E}_t\left[\min(r_t(\theta)\hat{A}_t,\mbox{clip}(r_t(\theta),1-\epsilon,1+\epsilon)\hat{A}_t)\right]
\end{split}
\end{equation}
The clipping ensures that the ratio $r_t(\theta)$ remains within the interval $[1-\epsilon,1+\epsilon]$. Choosing the minimum between the unclipped and clipped objectives ensures that the final objective serves as a lower bound for the unclipped objective.

In \emph{Adaptive KL Penalty Coefficient} (referred to as KLPEN in the notations), we apply the penalty on the KL divergence with adaptation of the penalty coefficient to achieve a destination value, $d_{\mbox{des}}$, for the KL divergence in each policy update:
\begin{equation} \small \label{Adaptive_KL}
\begin{split}
    L&^{KLPEN}(\theta)=\\&\hat{E}_t\left[\dfrac{\pi_\theta(\alpha_t|s_t)}{\pi_{\theta_{old}}(\alpha_t|s_t)}\hat{A}_t-\beta KL[\pi_{\theta_{old}}(\cdot|s_t),\pi_\theta(\cdot|s_t)]\right].
\end{split}
\end{equation}
The Adaptive KL Penalty Coefficient can be used instead or in addition to the Clipped Surrogate Objective. We denote:
\begin{equation} \small \label{L_KL}
    L^{KL}(\theta)= \hat{E}_t\left[-\beta KL[\pi_{\theta_{old}}(\cdot|s_t),\pi_\theta(\cdot|s_t)]\right]
\end{equation}
as the Adaptive KL Penalty from (\ref{Adaptive_KL}), and we formalize our PPO policy update objective using the two solutions mentioned above:
\begin{equation} \begin{split}\small \label{our_objective} L_t(\theta)&=L^{CLIP}+L^{KL}\\&=\hat{E}_t[\min(r_t(\theta)\hat{A}_t,clip(r_t(\theta),1-\epsilon,1+\epsilon)\hat{A}_t)\\&\hspace{0.5cm}-\beta KL[\pi_{\theta_{old}}(\cdot|s_t),\pi_\theta(\cdot|s_t)]], 
\end{split}
\end{equation}
where $\beta$ is changed by computing $d\triangleq \hat{E}_t[ KL[\pi_{\theta_{old}}(\cdot|s_t),\pi_\theta(\cdot|s_t)]]$, and dividing $\beta$ by a coefficient when $d<d_{\mbox{des}}/c_{\mbox{des}}$, or multiply $\beta$ by a coefficient when $d>d_{\mbox{des}}\cdot c_{\mbox{des}}$. Here, $c_{\mbox{des}}$ is an hyper-parameter. The initial value of $\beta$ is less important since it is quickly adjusted.

\begin{figure*} \includegraphics[width=\textwidth]{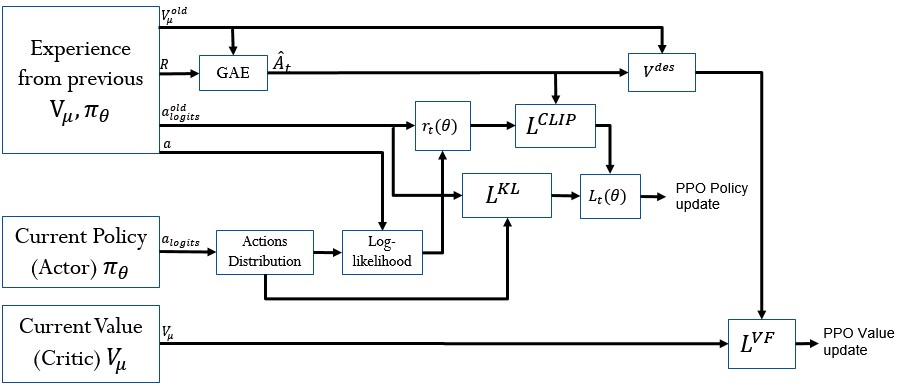}
    \caption{An illustration of the PPO Update in the MARLA algorithm.}
    \label{Figure:PPO_blocks}
\end{figure*}

For the Value network we use a squared-error loss:
\begin{equation} \label{vf_update}
    L^{VF}_t(\mu)=(V_{\mu}(s_t)-V^{\mbox{des}}_t)^2,
\end{equation}
where $V_{\mu}(s_t)$ is the Value network output, and
\begin{equation} \label{v_des}
    V^{\mbox{des}}_t=\hat{A}_t+V_{\mu_{old}}(s_t).
\end{equation}
Here, $V_{\mu_{old}}(s_t)$ is the previous Value network output. See Fig.\ref{Figure:PPO_blocks} for the block diagram of the algorithm.

\begin{figure*} \includegraphics[width=\textwidth]{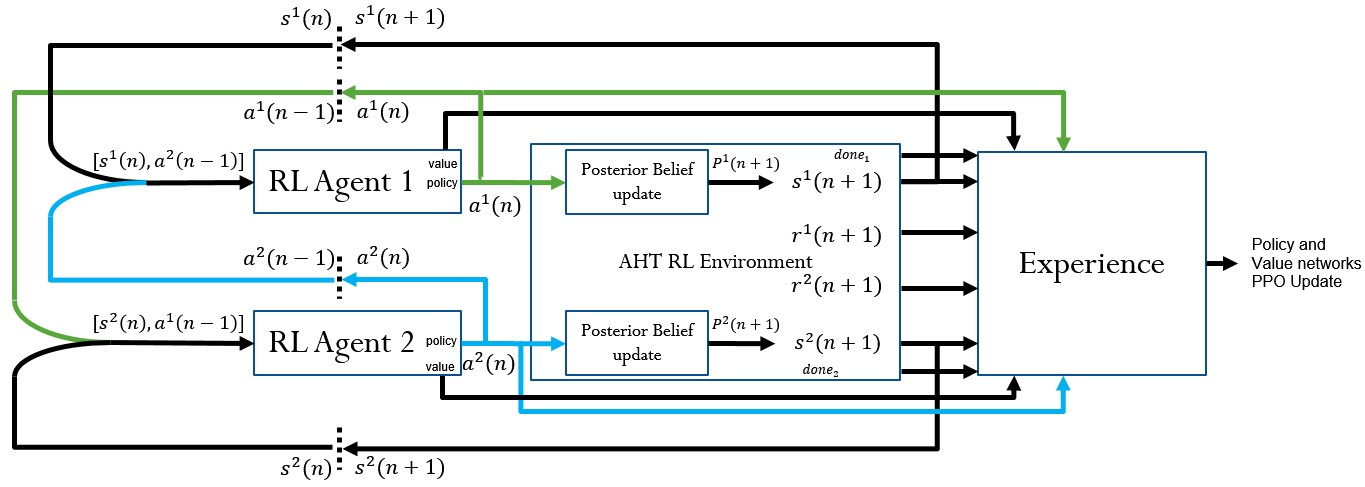}
    \caption{An illustration of the RL Episode in the MARLA algorithm.}
    \label{Figure:Episode_one_step_blocks}
\end{figure*}

\subsection{Multi-Agent Learning for Decentralized Active Hypothesis Testing}

\subsubsection{Agent Networks and Training} For the AHT task, MARLA adopts an Actor-Critic algorithm with PPO updates. Notably, it incorporates both clipping and an adaptive penalty coefficient into its methodology. An important characteristic of the MARLA framework is that the Policy and Value function networks do not share parameters between them. The learning process in MARLA follows a Centralized Training and Decentralized Execution (CTDE) structure. This structure offers the value network of each agent additional information regarding the dynamics of other agents. In the CTDE framework, each agent's value function network receives inputs comprising the state of its respective agent (which is the posterior belief regarding all hypotheses), the states of all other agents, and a communication signal that is influenced by past actions and observations. The actions taken by the agents in MARLA correspond to experiments or samples, which are instrumental in updating the posterior belief regarding the true hypothesis. A policy network with shared parameters is employed. This shared parameterization is essential because all agents must learn the same sampling behavior while possessing the ability to "listen" and effectively exploit the communication messages from other agents. To cope with the non-stationary POMDP environment, each agent is equipped with a fully observable critic. This configuration is designed to enhance the agent's ability to adapt to the dynamic nature of the environment.

\subsubsection{Agent Communication} It is essential to facilitate communication among the agents, as this communication serves the purpose of enabling agents to guide and assist one another in their decision-making processes. Importantly, this communication occurs through a rate-limited channel. The communication message, denoted as $m^k(n)$, serves to provide additional information to each agent about the dynamics of other agents. In this context, it is of utmost importance for agents to acquire the skill of determining when it is beneficial to "listen" to the decisions made by their fellow agents. This process plays a pivotal role in enhancing two important aspects: first, it aids in optimizing the selection of sampling actions to gather the most informative observations, and second, it contributes to improving the overall accuracy of the agents' declarations. This setup enables agents to effectively collaborate and make informed decisions, harnessing the information exchanged over the rate-limited communication channel. For a visual representation of the episode one-step block diagram, which includes the previous action as the communication message ($m^k(n)=a^k(n-1)$), please refer to Figure \ref{Figure:Episode_one_step_blocks}.

\subsection{Detailed Description of the MARLA Algorithm}

The pseudocode of the MARLA algorithm is provided in Algorithm 1. At the beginning of each episode, the posterior belief is set to a normal distribution, and a random selection is done for the episodes true hypothesis, $\mathbf{H}$. At each time $n$ of the episode, the agents select an experiment using the previous posterior belief each one holds, $\mathcal{P}^k(n-1)$, and the previous experiment done by all other agents, $m^{\bar{k}}(n)$. The experiment results with new observation, subjected to noise $\textbf{G}$, followed by a posterior belief $\mathcal{P}^k(n)$ update. The samples are stored for training the agents. Inactive agents share the communication message for $T'$ times and are seen by the fully observable critic as a zero state. Coupled reward is computed as episode ends. The agents are trained by SGD with multi-epoch learning. An open-source software implementation is available at \cite{szostak2023deep_code}.

\begin{figure}[ht]
\centering
\label{fig_2}
\includegraphics[width=8.5cm]{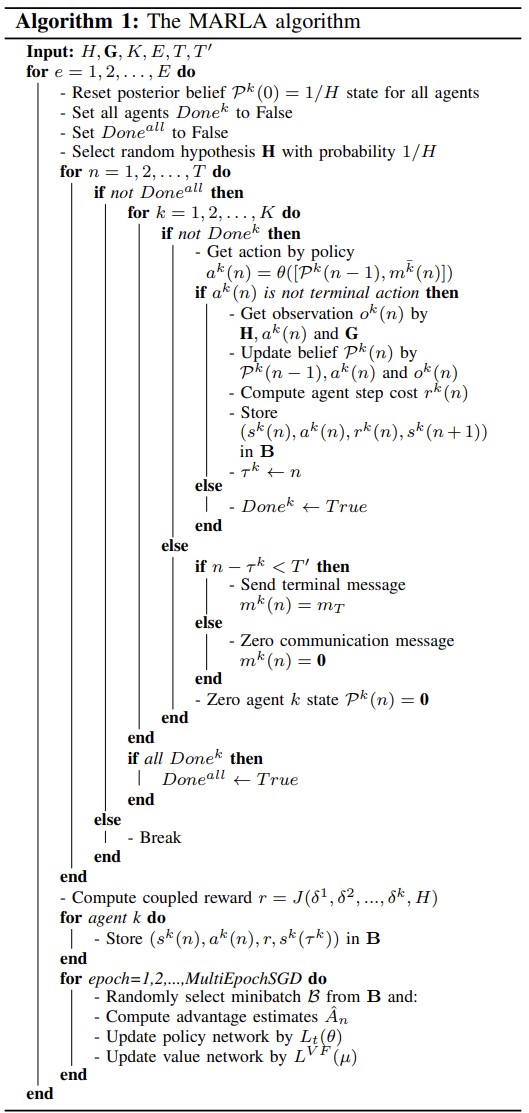}
\caption{Pseudocode of the MARLA algorithm.}
\end{figure}

\section{Experiments}
\label{sec:sim}

In this section we present experimental results of the MARLA algorithm for a decentralized active anomaly detection problem. We start by describing the simulation setting for this problem in Subsection \ref{ssec:sim_setting}, and then present the simulation results in Subsection \ref{ssec:sim_results}.

\subsection{Simulation Setting}
\label{ssec:sim_setting}

Active anomaly detection is an important application of AHT, which was studied extensively in recent years under various settings and assumptions \cite{lai2011quickest, caromi2012fast, tajer2013quick, cohen2014optimal, cohen2015active, cohen2015asymptotically, geng2016quickest, heydari2016quickest, kaspi2017searching, huang2018active, heydari2018quickest,  tsopelakos2019sequential, gurevich2019sequential, rosenthal2020arba, hemo2020searching, wang2020information, gafni2021searching, lambez2021anomaly, tsopelakos2022sequential, chaudhuri2022joint, kartik2022fixed}. 

In this specific task, the objective is to identify an anomalous process from within a collection of processes. The normal processes, when sampled, generate noisy observations drawn from a typical distribution that represents normal behavior. In our experiments, this distribution is simulated as Gaussian noise. On the other hand, an anomalous process, when sampled, yields noisy observations drawn from a distribution that deviates from the typical distribution. In our experiments, this anomalous distribution is simulated as a DC signal plus Gaussian noise. Each agent has the ability to control the observation distribution by selecting which process to sample at any given time. It is worth noting that calculating a structured optimal sampling strategy for this problem is a mathematically intractable task, even in the single-agent setting. In such cases, only asymptotically optimal solutions have been developed, as discussed in \cite{chernoff1959sequential}. In contrast, the MARLA algorithm we propose addresses this challenge by optimizing the Bayes risk associated with the problem using the DRL optimization method. This approach provides a practical means to tackle the problem and achieve effective solutions.

The proposed MARLA framework was developed using Python and is available at GitHub (see link in \cite{szostak2023deep_code}). The implementation is based on Ray \cite{liang2017ray} Rllib \cite{liang2018rllib} library, which is a scalable DRL 
library designed for both single and multi-agent settings. We designed and developed a custom-made multi-agent AHT environment for the active anomaly detection task based on OpenAI Gym API \cite{brockman2016openai} for integration with Ray Rllib library. We used Rllib's PPO implementations and designed our Policy and Value networks architecture for the MARLA framework. The key design features are described below.

Communication among agents is facilitated through short-message signals, which play a pivotal role in their collaboration and the accomplishment of the detection task. In our simulations, we utilize information related to the selected action as the basis for these short-message signals. The rationale behind this choice is that agents can enhance the reliability of their inferences concerning the location of the anomalous process by considering the actions chosen by other agents. This aspect is particularly intriguing because it demonstrates that agents can acquire valuable insights from the actions chosen by their peers without the requirement for transmitting explicit observations, which typically demands significantly higher bandwidth. Furthermore, it is important to note that when full communication of observations occurs among agents, the problem effectively reduces to a single-agent scenario with multiple samplers. Thus, the ability to learn solely from the actions of other agents underscores the true power of the multi-agent setting. When an agent decides to conclude its testing, it shares its decision, whether it is correct or incorrect, concerning the anomaly's location to all other agents. Note that significant errors can arise in situations where an agent is unable to deduce the location of the anomalous process based on its observation distribution. This highlights the challenge posed by situations where the observations provide insufficient information for precise inference. Thus, collaboration among agents becomes even more crucial as it can significantly enhance overall performance by pooling together their collective inference capabilities, compensating for the limited information available from their individual observations. The simulation results we present illustrate that agents can effectively learn how to collaborate using these short-message signals. Moreover, these simulations showcase the substantial improvement in their detection performance achieved through the use of the MARLA algorithm.

At each time step $n$, every agent acquires its state from the environment. This state encompasses the agents' beliefs concerning the hypotheses, specifically the dynamically updated posterior probabilities determined by: 
\begin{align}
\mathcal{P}_j(n+1)=
\begin{cases} \label{anomaly_detection_belief_update}
    \dfrac{\mathcal{P}_j(n)f(o^k(n))}{p(o^k(n)|i(n))}, & \text{if } j\not=i(n),\vspace{0.2cm}\\
    \dfrac{\mathcal{P}_j(n)g(o^k(n))}{p(o^k(n)|i(n))}, & \text{if } j=i(n),
\end{cases}
\end{align}
where $i(n)$ is the action selected by the agent at time $n$, $o^k(n)$ is the collected observation, $f$ is the normal probability density, $g$ is the abnormal probability density and
\begin{equation} \begin{split}
    p(o^k(n)&|i(n))=\\&(1-\mathcal{P}_{i(n)}(n))f(o^k(n))-\mathcal{P}_{i(n)}(n)g(o^k(n)).
\end{split}
\end{equation}
Additionally, the state includes the short-message signals received from other agents. The training process was carried out in a centralized manner, after which the trained neural network (i.e., trained actor) was employed by each agent in a decentralized fashion during the online implementation. This approach ensures consistency and coordinated decision-making among the agents based on their shared training.

We utilized a Fully-Observable Value network (Centralized Critic) for each agent in conjunction with a Policy network (Actor) that featured shared parameters. Our Value network received input comprising the current agent's observation and the observations and actions of all other agents from the previous step. Meanwhile, the Policy network was fed with the current agent's observation and the previous actions of all other agents. We tested the algorithm in different environment configurations, including: (i) \emph{Independent Agents:} In this scenario, each agent had the capability to sample from all processes. Consequently, the action space was as extensive as the number of processes, with an additional action for the termination of the test. (ii) \emph{Agents with Overlap:} In this case, each agent could sample a subset of the processes, and some processes were eligible to be sampled by multiple agents. The action space was determined by the portion of processes that each agent could sample, with an additional action for the termination of the test. (iii) \emph{Agents without Overlap:} Here, each process could be sampled by only one agent exclusively. The action space was defined as the total number of processes divided by the total number of agents, with an additional action for the termination of the test. These distinct environment options allowed us to assess the performance of the MARLA algorithm under various conditions and agent configurations.

\subsection{Results}
\label{ssec:sim_results}
\subsubsection{Independent Agents}
We simulated scenarios with $5, 7$, and $10$ processes, each containing one anomalous process. In these experiments, we deployed two agents that collaborated via communications to identify the anomalous process effectively. In addition to this collaborative setting, we also examined a scenario where the two agents did not engage in communication, serving as a performance benchmark. This comparison allowed us to demonstrate the capability of the agents to learn and enhance their collaborative performance through the MARLA algorithm. 

\begin{figure*}  
    \centering
    \begin{subfigure}[b]{0.49\textwidth}
        \includegraphics[width=\textwidth]{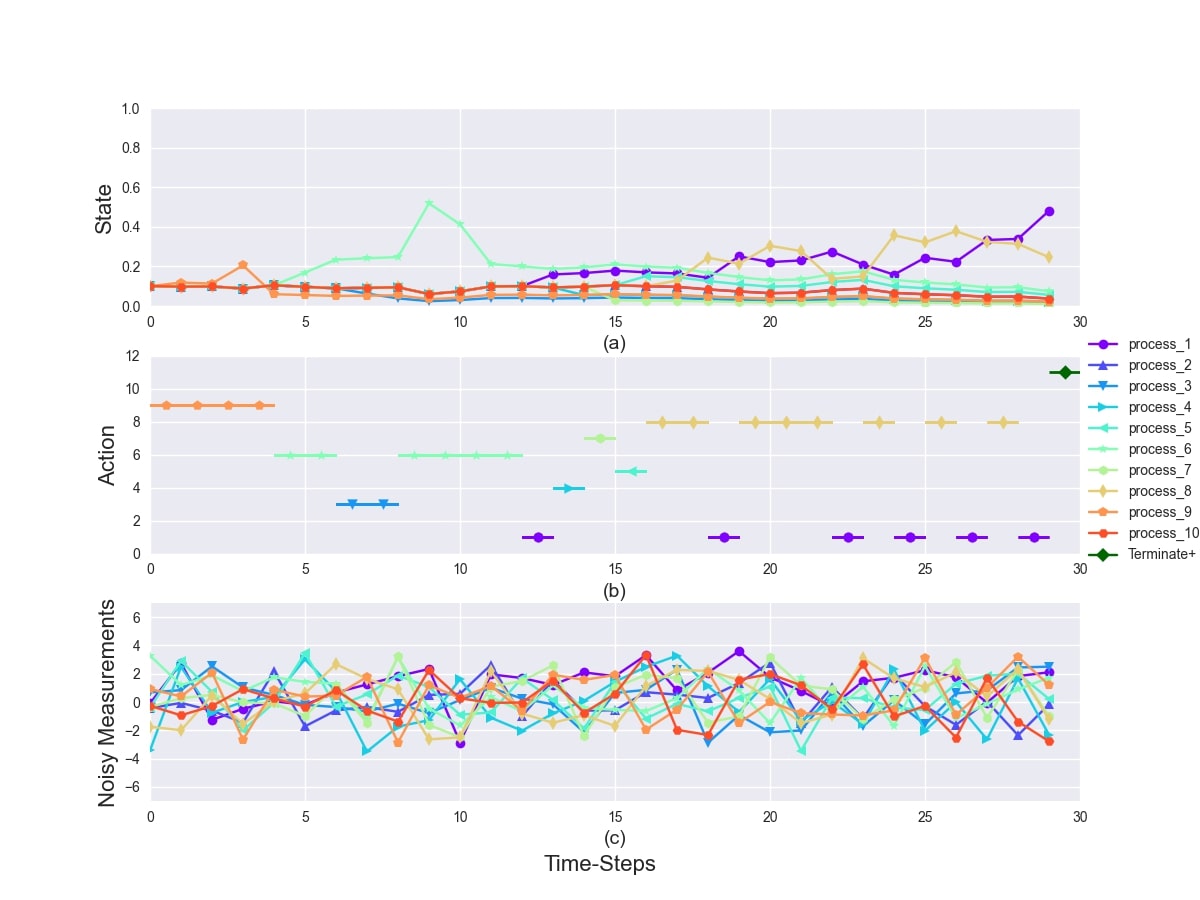}
        \label{figure:10_agent_0_independent_agents}
    \end{subfigure}
    \hfill
    \begin{subfigure}[b]{0.49\textwidth}
        \includegraphics[width=\textwidth]{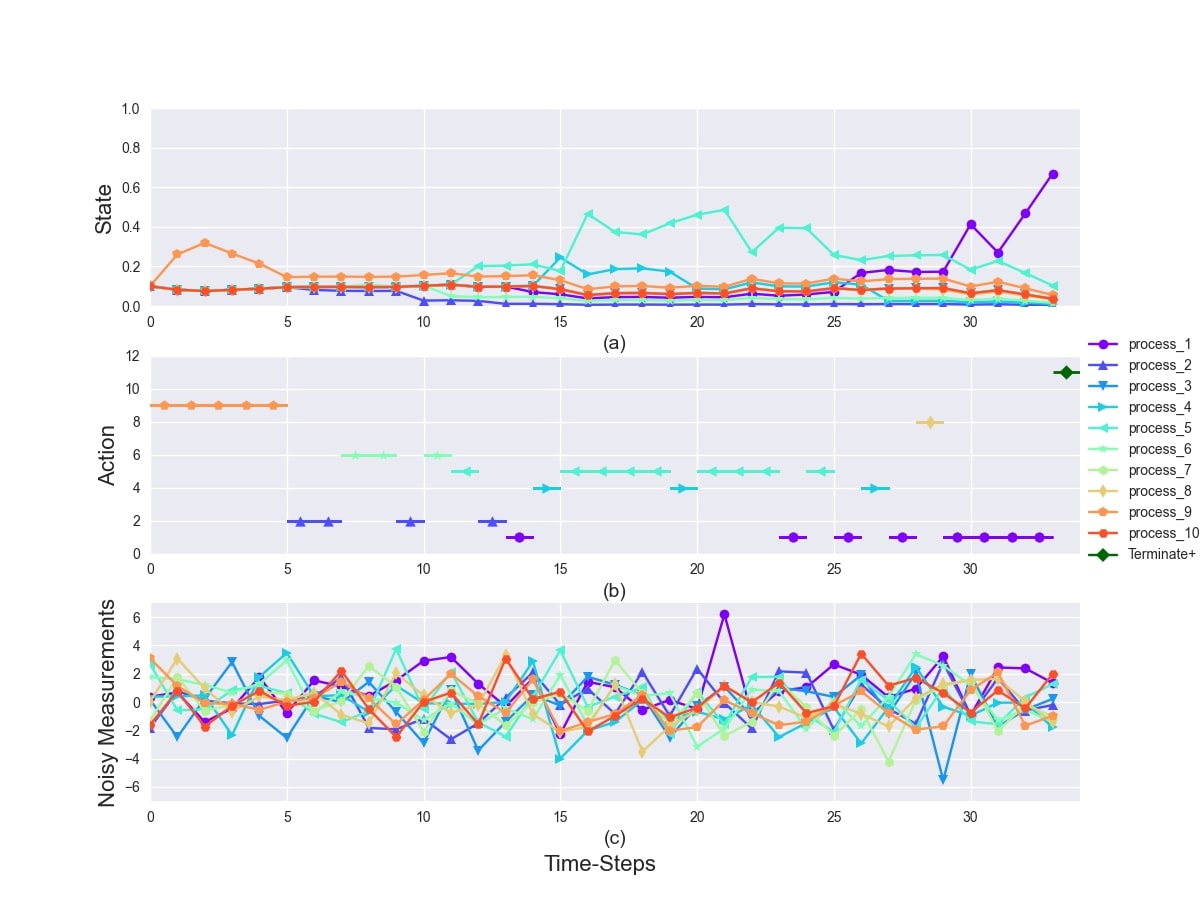}
        \label{figure:10_agent_1_independent_agents}
    \end{subfigure}
    \caption{Simulation results for an episode with $10$ processes featuring independent agents: (a) depicts the posterior belief regarding the hypotheses (State), (b) illustrates the actions taken at each time step, and (c) shows the noisy measurements. The anomalous process (i.e., the true hypothesis) is located in cell 1.}
    \label{Figure:independent_agents}
\end{figure*}

\begin{figure}[ht]
\centering
\includegraphics[width=7cm]{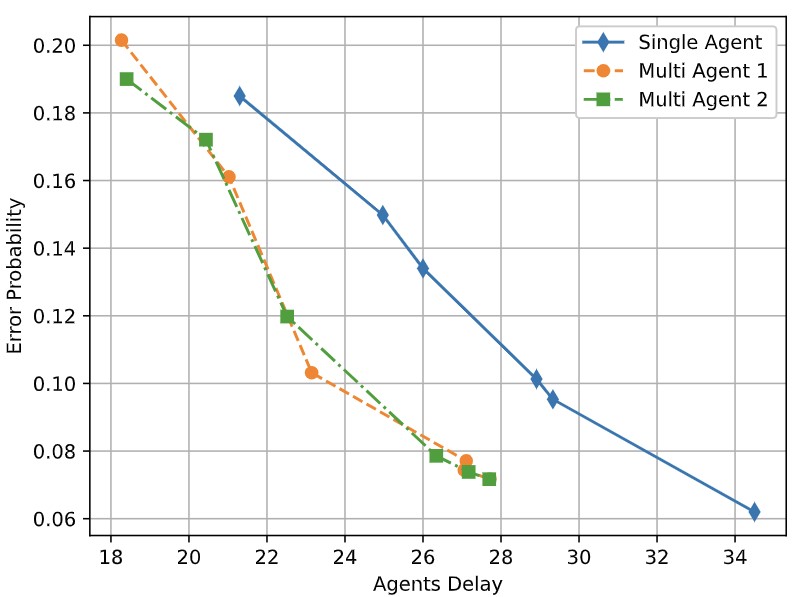}
\caption{The empirical error rate for $5$ processes featuring independent agents.}
\label{Figure:5_processes_error_rate}
\end{figure}

\begin{figure}[ht]
\centering
\includegraphics[width=7cm]{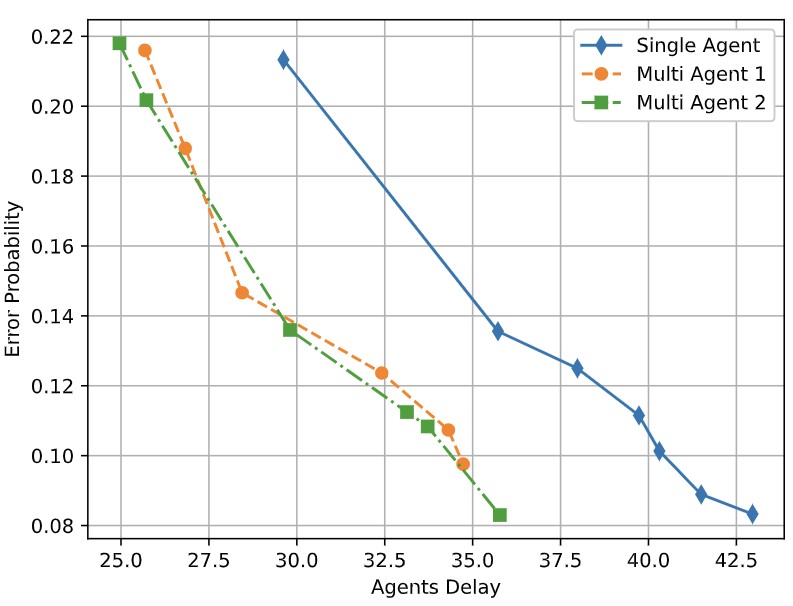}
\caption{The empirical error rate for $7$ processes featuring independent agents.}
\label{Figure:7_processes_error_rate}
\end{figure}

\begin{figure}[ht]
\centering
\includegraphics[width=7cm]{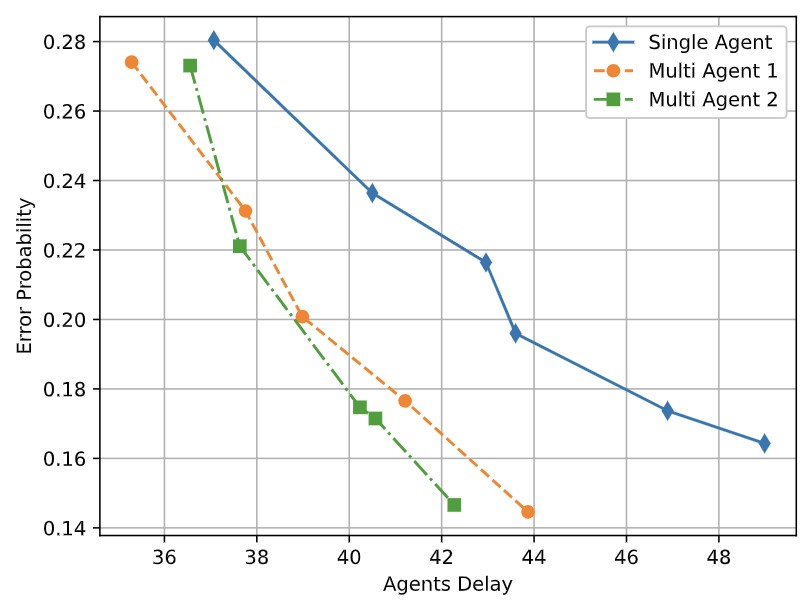}
\caption{The empirical error rate for $10$ processes featuring independent agents.}
\label{Figure:10_processes_error_rate}
\end{figure}

We commence by illustrating the learning process of the two agents as they progress toward the optimal policy trained through DRL optimization. We use an example of one episode to demonstrate agents' dynamics in a scenario with $10$ processes, focusing on a case where the agents make correct decisions to highlight policy dynamics. Error rates will be discussed and presented in subsequent figures (Figs. \ref{Figure:5_processes_error_rate}, \ref{Figure:7_processes_error_rate}, and \ref{Figure:10_processes_error_rate}). In the first sub-figure (\ref{Figure:independent_agents}.a), we depict the posterior probabilities, often referred to as the belief regarding the location of the anomalous process (i.e., the states in the MARLA framework). As time progresses, it becomes evident that the belief in the true process approaches one. This indicates that the agents effectively collaborate to infer the actual location of the process. In the second sub-figure (\ref{Figure:independent_agents}.b), we present the selected processes that the agents choose to sample over time (i.e., the selected actions in the MARLA framework). The noisy measurements are presented in the third sub-figure (\ref{Figure:independent_agents}.c). It is noticeable that locations with higher beliefs are sampled more frequently, and the sampling of the true location becomes more frequent over time as the decision becomes more accurate. This dynamic aligns with the asymptotically optimal policy developed in \cite{cohen2015active} for infinite time horizons. This policy dictates that as time progresses, the optimal policy converges to one that selects the process with the highest belief (which is the true location with high probability, as the error probability approaches zero with increasing detection time). 

\begin{figure}[ht]
\centering
\includegraphics[width=7cm]{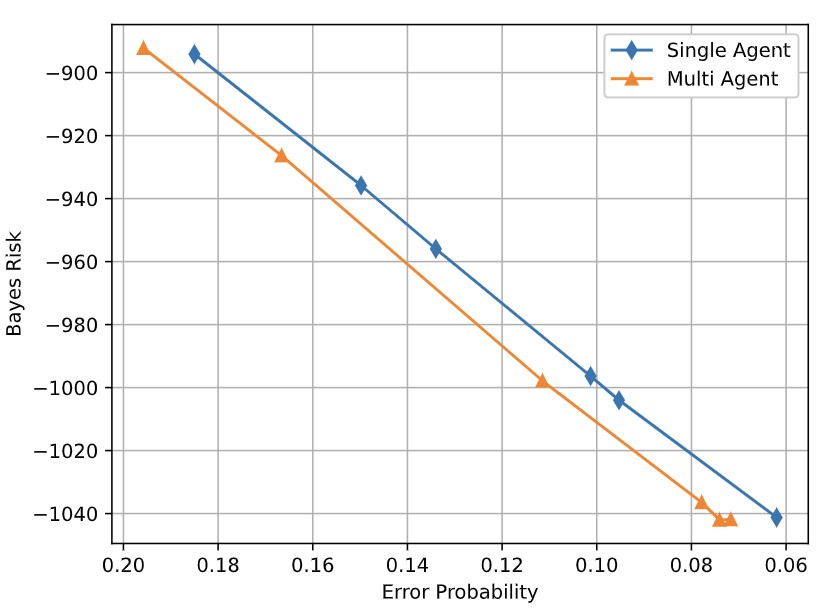}
\caption{The empirical Bayes risk for $5$ processes featuring independent agents.}
\label{Figure:5_processes_bayes_risk}
\end{figure}

\begin{figure}[ht]
\centering
\includegraphics[width=7cm]{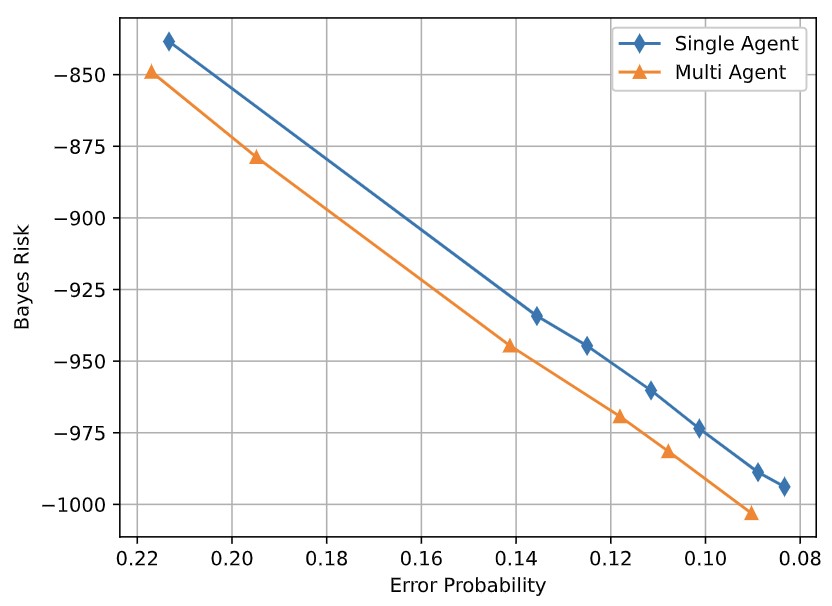}
\caption{The empirical Bayes risk for $7$ processes featuring independent agents.}
\label{Figure:7_processes_bayes_risk}
\end{figure}

\begin{figure}[ht]
\centering
\includegraphics[width=7cm]{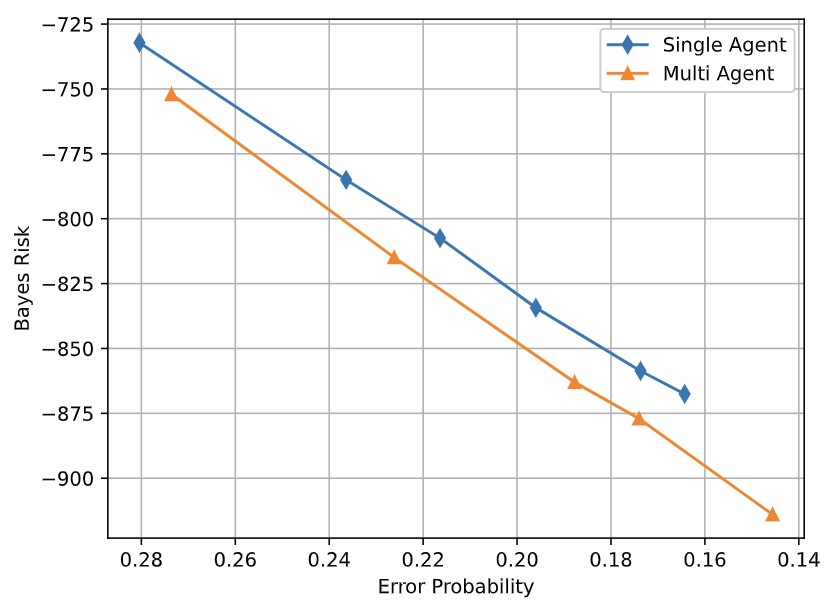}
\caption{The empirical Bayes risk for $10$ processes featuring independent agents.}
\label{Figure:10_processes_bayes_risk}
\end{figure}

Next, we present the empirical error probability in relation to the average sample size, which is a measure of the average detection delay. These results are depicted in Figs.\ref{Figure:5_processes_error_rate}, \ref{Figure:7_processes_error_rate}, and \ref{Figure:10_processes_error_rate} for scenarios involving 5, 7, and 10 processes, respectively. In addition to the multi-agent setting, we conducted experiments in a single-agent setting as well as a benchmark for performance assessment. It is evident from the figures that agent collaboration leads to a reduction in the detection delay for a given error probability value, with a reduction of approximately $15\%$ to $20\%$ compared to the single-agent setting. Moreover, substantial reductions in terms of Bayes risk are observed, as demonstrated in Figs. \ref{Figure:5_processes_bayes_risk}, \ref{Figure:7_processes_bayes_risk}, and \ref{Figure:10_processes_bayes_risk}.

\subsubsection{Agents with Overlap} 
\begin{figure}[ht]
\centering
\includegraphics[width=7cm]{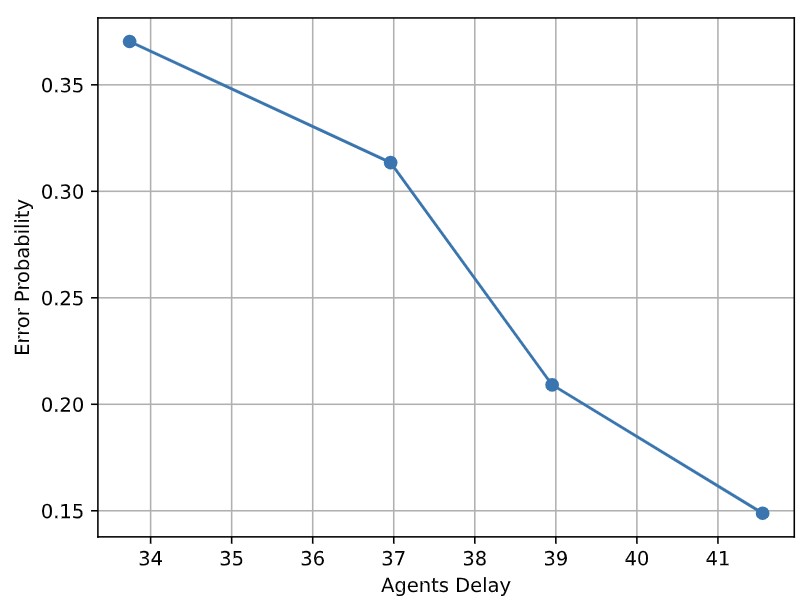}
\caption{The empirical error rate (averaged over two agents) for $10$ processes featuring agents with overlap.}
\label{Figure:10_processes_with_overlap_error_rate}
\end{figure}

\begin{figure}[ht]
\centering
\includegraphics[width=7cm]{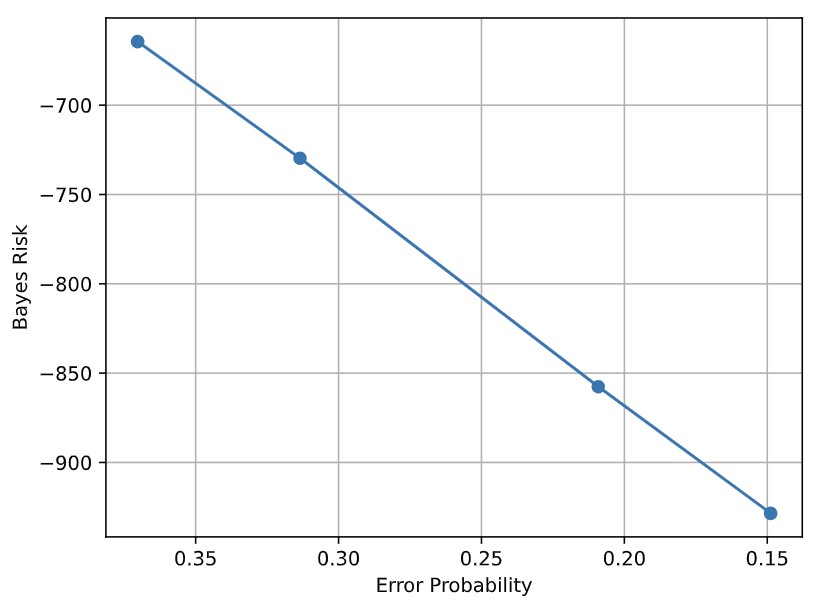}
\caption{The empirical Bayes risk (averaged over two agents) for $10$ processes featuring agents with overlap.}
\label{Figure:10_processes_with_overlap_bayes_risk}
\end{figure}

For agents with overlap, we conducted experiments involving $10$ processes, one of which was anomalous. Two agents were utilized in these experiments, as previously described. Agent $1$ had the ability to sample a subgroup of processes numbered $1$ to $6$, while Agent $2$ could sample a subgroup of processes numbered $5$ to $10$. The error rate and Bayes risk are presented in Fig. \ref{Figure:10_processes_with_overlap_error_rate} and Fig. \ref{Figure:10_processes_with_overlap_bayes_risk}, respectively. This scenario is particularly interesting, as it is possible that one agent cannot accurately infer the true hypothesis without collaboration even with a large number of observations, if the true hypothesis is outside its sampling area. Nevertheless, it is evident that both agents efficiently learned to collaborate in identifying the anomalous process.

\begin{figure}[ht]
\centering
\includegraphics[width=7cm]{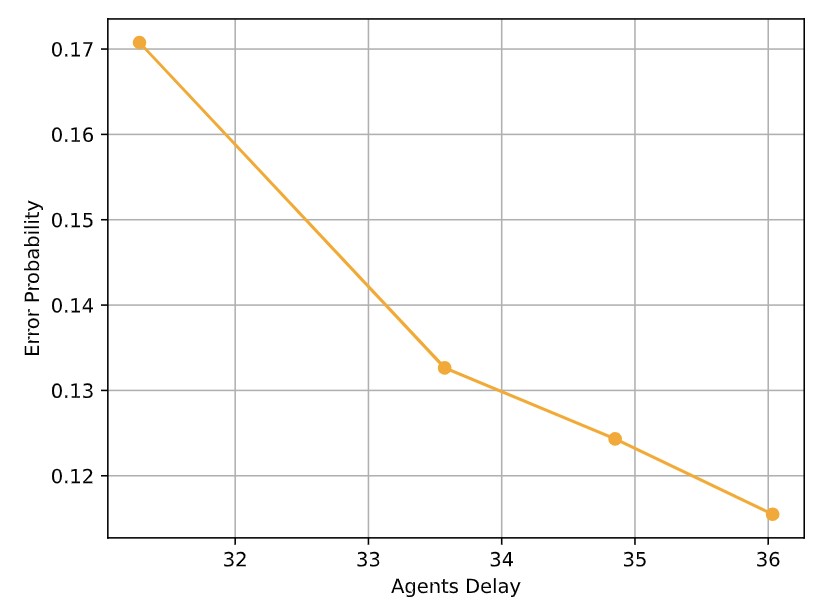}
\caption{The empirical error rate (averaged over two agents) for $10$ processes featuring agents without overlap.}
\label{Figure:10_processes_without_overlap_error_rate}
\end{figure}

\begin{figure}[ht]
\centering
\includegraphics[width=7cm]{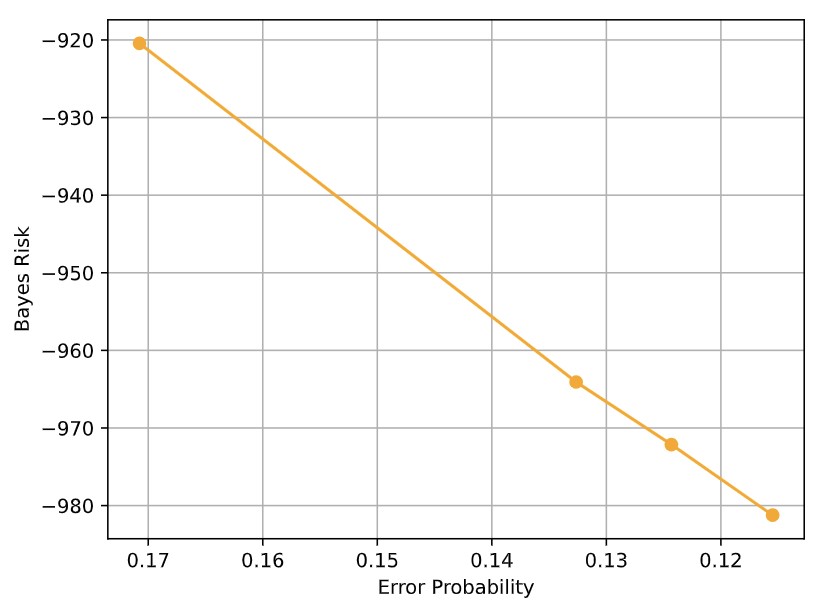}
\caption{The empirical Bayes risk (averaged over two agents) for $10$ processes featuring agents without overlap.}
\label{Figure:10_processes_without_overlap_bayes_risk}
\end{figure}

\subsubsection{Agents without Overlap} For agents without overlap, we conducted experiments involving $10$ processes, one of which was anomalous. Two agents were utilized in these experiments, as previously described. Agent 1 had the ability to sample a subgroup of processes 1 to 5, and Agent 2 could sample a subgroup of processes 6 to 10. The error rates and Bayes risk are presented in Fig. \ref{Figure:10_processes_without_overlap_error_rate} and Fig. \ref{Figure:10_processes_without_overlap_bayes_risk}, respectively. This scenario is particularly interesting because in each experiment in this case, there is always one agent that cannot accurately infer the true hypothesis, even with a large number of observations. This is due to the absence of overlap, and thus the true hypothesis is always outside one agent's sampling area. Nevertheless, it is evident that both agents efficiently learned to collaborate in identifying the anomalous process.

\section{Conclusion}

In this paper, we have formulated and explored a decentralized AHT problem, in which multiple agents collect noisy observations from the environment and must infer the true hypothesis from a set of hypotheses. These agents control observation distributions through sampling actions and collaborate to achieve the AHT task. Communication between agents is facilitated over a rate-limited channel. Our objective has been to design a multi-agent policy that minimizes the Bayes risk, encompassing both the sampling cost and the collective terminal cost incurred when declaring a hypothesis. To tackle this challenge, we introduced a novel algorithm grounded in a deep multi-agent reinforcement learning framework, dubbed MARLA. At each time step, each agent leverages a trained deep neural network to map its state to an action, with the aim of minimizing the Bayes risk. Our experimental results offer compelling evidence that agents can effectively learn to collaborate and enhance their performance through the use of MARLA. Furthermore, these results demonstrate the superiority of MARLA over single-agent learning. To foster further research and development in related fields, we have also made the MARLA framework available as open-source software.

\end{document}